# Simple Inverse Kinematics Computation Considering Joint Motion Efficiency

Ansei Yonezawa, *Member, IEEE*, Heisei Yonezawa*, Member, IEEE*, and Itsuro Kajiwara

*Abstract*— **Inverse kinematics is an important and challenging problem in the operation of industrial manipulators. This study proposes a simple inverse kinematics calculation scheme for an industrial serial manipulator. The proposed technique can calculate appropriate values of the joint variables to realize the desired end-effector position and orientation while considering the motion costs of each joint. Two scalar functions are defined for the joint variables: one is to evaluate the end-effector position and orientation, whereas the other is to evaluate the motion efficiency of the joints. By combining the two scalar functions, the inverse kinematics calculation of the manipulator is formulated as a numerical optimization problem. Furthermore, a simple algorithm for solving the inverse kinematics via the aforementioned optimization is constructed on the basis of the simultaneous perturbation stochastic approximation with a norm-limited update vector (NLSPSA). The proposed scheme considers not only the accuracy of the position and orientation of the end-effector but also the efficiency of the robot movement. Therefore, it yields a practical result of the inverse problem. Moreover, the proposed algorithm is simple and easy to implement owing to the high calculation efficiency of NLSPSA. Finally, the effectiveness of the proposed method is verified through numerical examples using a redundant manipulator.**

## I. INTRODUCTION

### A. Motivation

**R**OBOTS are indispensable to most industrial fields at present [1], [2]. The inverse kinematics (IK) problem must be solved to perform a desired task using a robot. Robot teaching [3] conducted by an operator is a typical and common method for solving the IK problem because it is equivalent to obtaining the proper joint configuration in order to realize the desired position and orientation of the end-effector. However, such teaching is expensive, tedious, and even dangerous in many situations. In particular, in the

Manuscript received Month xx, 2xxx; revised Month xx, xxxx; accepted Month x, xxxx. This work was supported in part by JSPS KAKENHI Grant Numbers JP22KJ0066, JP23K19084. *(Corresponding author: Ansei Yonezawa).*

A. Yonezawa, H. Yonezawa, and I. Kajiwara are with Division of Mechanical and Aerospace Engineering, Hokkaido University, Sapporo, Hokkaido, 060-8628, Japan (e-mail: ayonezawa[at]eng.hokudai.ac.jp; yonezawah[at]eng.hokudai.ac.jp; ikajiwara[at]eng.hokudai.ac.jp).

Color versions of one or more of the figures in this article are available online at http://ieeexplore.ieee.org

ongoing industrial revolution, referred to as Industry 4.0 [4], [5], manufacturing systems are more and more flexible than traditional systems (e.g., introduction of mobile manipulators [6], smart production-logistics system based on cyber-physical system [7]) in order to meet the requirements of individual customers [8], [9]. In other words, products and work descriptions are frequently changed depending on customers' demands. There are numerous challenges in implementing Industry 4.0 [10], among which changing robot tasks frequently implies that manual robot teaching is a critical bottleneck in making a manufacturing system flexible. Industrial robots must adapt autonomously to each task. Therefore, a simple and practical solving technique for the IK problem is the key to realizing not only efficient on-site robot operation but also flexible and autonomous manufacturing systems.

Compared with forward kinematics (FK), IK calculation is more challenging because the closed-form solutions cannot be obtained in general, and multiple solutions may exist [11]. Therefore, an IK calculation scheme is one of the most important technologies for improving the production capability and safety in manufacturing. In other words, developing a numerical method for IK is indispensable for making robots, including serial industrial manipulators, behave as desired.

It should be noted that industrial manipulators must move as compactly as possible in order to improve work efficiency and reduce energy consumption. Multiple solutions of numerical IK calculation are not equally desirable in the sense that the joint movement amounts given by each solution may be different; some of them will yield inefficient motion. Consequently, IK calculation must consider the motion cost of a manipulator.

### B. Related Work: IK Computation

Numerous studies have focused on solving the inverse calculation problem owing to its importance for robot operations [11]–[13]; specifically, several researches have focused on IK calculation for redundant robots [14]–[18]. Obtaining an analytical solution directly is a typical method for solving the IK problem. Many methods have been examined, because such solutions are generally the most accurate solutions [19]. For example, the IK mapping of serial robots with a spherical wrist has been derived on the basis of conformal geometric algebra [20], and an analytical IK computation for a rail-type redundant manipulator has been developed [21]. However, closed-form solutions do not always exist depending on the geometry of the robot [22].

Many studies have developed numerical techniques for the IK problem instead of deriving analytical solutions. The gradient-based method is a traditional method for numerical IK calculation (a comprehensive review of this approach can be found in [23]). A typical approach is the application of the Newton–Raphson method for nonlinear simultaneous equations $e({}^{curr}r, {}^*r) = 0$, defined by errors $e$ between the current state ${}^{curr}r$ and the desired state ${}^*r$ of the end-effector (${}^{curr}r$ is determined by the current joint configuration) [24]. This technique often fails depending on the problem setting [23], [25]. Another strategy of such a method is minimizing a scalar function $e^T({}^{curr}r, {}^*r)Me({}^{curr}r, {}^*r)$ instead of directly solving $e({}^{curr}r, {}^*r) = 0$, where $M$ denotes a constant weighting matrix (e.g., the Newton–Raphson method, the Levenberg–Marquardt method) [23]. However, most of these gradient-based IK calculations require inverse matrix computations, which make the algorithm complicated. Moreover, in gradient-based methods that require inverse computations, the IK calculation often fails when the manipulator is nearly singular because of numerical instability.

Learning methods such as approximating the IK function via neural networks are strong techniques for solving the IK problem. The IK problem for a three-link robotic manipulator has been addressed via a feed-forward neural network [26]. Furthermore, an unsupervised-learning algorithm has been developed for solving the IK problem of a mobile continuum manipulator [27]. In addition, a mapping function of FK and its Jacobian for the IK computation have been obtained through neural networks [28]. Moreover, an IK calculation scheme for a redundant manipulator has been developed using the growing neural gas network [29]. However, these methods must be trained for the geometry and task of each manipulator, which decreases the versatility. Learning for various situations requires a long training time and high development costs.

Recently, IK computations based on metaheuristics, such as evolutionary algorithms and swarm intelligence, have been widely studied [30]. These techniques are powerful because of their ability to search for solutions as well as their flexibility. Regarding evolutionary algorithms, the genetic algorithm (GA) is a predominant technique for solving the IK problem. One study has focused on solving the IK problem of a serial manipulator via an adaptive niching GA [31]. The GA has been frequently combined with learning-based IK algorithms. For example, an IK solution preliminary computed via an extreme learning machine has been subsequently optimized through a sequential mutation GA [32]. In addition, novel evolutionary algorithms have been developed for solving the IK problem [19], [33]. As for swarm-based algorithms, the particle swarm optimization (PSO) method and its variants, such as improved PSO [34] and bidirectional PSO [35], have been widely studied. One study has proposed the modified version of fully resampled PSO for the IK problem [36]. Nevertheless, evolutionary methods and swarm techniques involve complex algorithms and tedious hyperparameter tuning, which adversely affect their practical application.

*Remark 1:* As mentioned above, the IK problem generally has multiple solutions. Users must select an appropriate solution from the multiple possible solutions because not all the solutions are always equally well suited to real-world applications; some of them may cause undesirable and excessive joint motion of the manipulator. The selection of the appropriate IK solution is a significant challenge in on-site robot operations. Consequently, an IK calculation method must be constructed such that the algorithm can provide a practically feasible solution considering motion efficiency. To the best of our knowledge, there are very few IK computation methods that consider the practicability of the obtained solution in terms of motion efficiency.

*Remark 2:* The IK problem is also found in the field of biological cybernetics, such as the analysis and modeling of human motion [37]. Specifically, the IK problem of the human posture can be formulated as an optimization problem of a measure expressing the efficiency of movement so as to explain the skilled motion of humans (e.g., minimization of jerk [38], and metabolic energy [39]). This perspective facilitates the selection of one solution from numerous candidate solutions in the IK problem of general manipulators, resulting in efficient robot operation. Inspired by these studies, this study develops a numerical IK calculation scheme from the viewpoint of motion cost minimization.

### C. Contribution and Novelty of the Study

A simple IK computation strategy is proposed for a general serial manipulator, considering motion efficiency. The IK problem for the manipulator is formulated as the minimization of the joint movement cost, where realizing the desired position and orientation of the end-effector is a constraint in the optimization problem. The constrained optimization problem is converted into an equivalent unconstrained problem. Then, the unconstrained problem is solved using the simultaneous perturbation stochastic approximation with a norm-limited update vector (NLSPSA). We employ NLSPSA owing to its high calculation efficiency, algorithmic simplicity, ability to search for a solution, and calculation stability. In particular, compared with previous studies, the contributions and novelties of this study can be summarized as follows:

1) (*Simple implementation*) The proposed method is simpler to implement than conventional gradient methods, since the proposed technique requires only FK mapping. Unlike traditional IK calculation schemes based on gradients [23]–[25], the proposed algorithm does not require analytical forms of the Jacobian or Hessian owing to the stochastically approximated gradient for NLSPSA. Moreover, the proposed technique has a simpler algorithm structure compared with traditional metaheuristics-based methods such as the GA [31], [32] and PSO [34]–[36]. In addition, expensive and tedious learning procedures are not required for the proposed scheme, which is a significant advantage over learning-based methods (e.g., [26]). Learning for specific manipulator and task degrades the versatility. Therefore, the proposed strategy can be easily implemented and is also able to flexibly handle various manipulators with minimal modifications. Simplicity and flexibility are significant advantages of the proposed method over traditional methods such as analytical gradient-, metaheuristics-, and learning-based schemes.

2) (*Calculation efficiency*) Unlike existing IK computations based on population-based metaheuristics (e.g., the GA [31], [32], PSO [34]–[36], the proposed algorithm involves less wasteful calculations because it is designed as a single-path search. Furthermore, owing to NLSPSA, the proposed strategy has high calculation efficiency in the sense that the algorithm requires only two FK calculations per iteration, regardless of the manipulator's degree of freedom. Thus, the proposed method is more computationally efficient than conventional IK computations based on population-based metaheuristics. In addition, explicit calculations of the Jacobian, Hessian, and their inverses are not necessary. Therefore, this algorithm is efficient and is especially well suited to redundant manipulators, which have received a great deal of attention owing to its flexibility [40], [41]. In this study, the efficiency of the NLSPSA-based computation is numerically demonstrated by comparing it with the PSO-based optimization as a conventional method.

3) (*Calculation stability*) The proposed computation strategy does not require inverse matrix calculations. Therefore, the proposed method achieves stable computations compared with conventional gradient-based methods and pseudoinverse methods.

4) (*Practicability of solutions*) This study is the first attempt to explicitly consider the practicability of solutions during the IK computation for a general serial manipulator, to the best of our knowledge. As mentioned in Remark 2, the proposed method explicitly evaluates the motion cost of the manipulator similar to the IK analysis in the field of biological cybernetics. This feature realizes small and efficient joint motions, thereby avoiding impractical solutions that yield awkward and excessive motions.

5) (*Impact on industry*) As discussed above, the proposed technique is simple and practical; hence, it drastically condenses the operation procedure (especially robot teaching, which is not only tedious but also dangerous) of industrial robots. The widespread use of this scheme will enhance the flexibility and safety of manufacturing systems, and thus promote the realization of Industry 4.0.

### D. Structure of the Paper

The remainder of this paper is organized as follows. Section II states the problem considered in this study and formulates the IK computation problem considering the joint motion cost. Section III presents an overview of NLSPSA and describes the proposed IK computation strategy. Section IV describes the application of the proposed method to IK computations of redundant manipulators in order to verify its validity. Section V discusses this approach as well as the validation results. Finally, Section VI concludes the paper.

### E. Notation

The sets of real numbers, strictly positive real numbers, and integers are denoted by $\mathbb{R}$, $\mathbb{R}_+$, and $\mathbb{Z}$, respectively. The set of nonnegative real numbers are represented as $\mathbb{R}_+ \cup \{0\}$. The symbol $\mathbb{R}^p$ denotes the set of $p$-dimensional real vectors.

## II. PROBLEM STATEMENT

### A. Forward and Inverse Kinematics

In this study, we consider a serial manipulator with $n$ degree of freedom ($n$-DOF serial manipulator). Throughout this study, the self-collision and obstacles are not considered. The outline of the IK problems is shown in Fig. 1. Let $\mathcal{C}$ be the configuration space of the manipulator, $q = [q_1 \quad q_2 \quad \cdots \quad q_n]^T \in \mathcal{C}$ be its joint configuration vector, $\mathcal{W}$ be the workspace (end-effector space), and $f: \mathcal{C} \to \mathcal{W}$ be the FK mapping determined by the manipulator geometry. The FK problem is the calculation of the position and orientation, denoted by $r \in \mathcal{W}$, of the end-effector from the given $q$ and $f$ as follows:

$$r = f(q). \tag{1}$$

Meanwhile, the IK problem is the calculation of the joint variable vector $q$ for realizing the desired position and orientation of the end-effector as follows:

$$q = f^{-1}(r). \tag{2}$$

In other words, for the desired position and orientation of the end-effector $r^*$, the IK problem is the computation of $q$ to satisfy

$$\epsilon(q, r^*) := r^* - f(q) = 0. \tag{3}$$

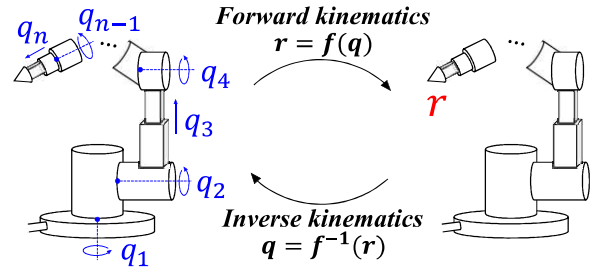

**Fig. 1.** Forward and inverse kinematics.

*Remark 3*: As discussed in Section I-B, it is difficult to obtain the analytical solution of (3) in general because this is a high-dimensional and nonlinear simultaneous equation. A traditional approach to numerically solving (3) is the Newton–Raphson method for a simultaneous equation $\epsilon(q, r^*) = 0$, which is a typical gradient-based IK computation technique. Nevertheless, this strategy often fails depending on the problem setting [23].

*Remark 4*: An alternative approach to solving (3) approximately is to solve the following optimization problem:

$$\underset{q}{\text{minimize}} \; J_{EE}(q)$$

$$J_{EE}(q) := \epsilon^T(q, r^*) R_{EE} \epsilon(q, r^*) \tag{4}$$
$$= \{r^* - f(q)\}^T R_{EE} \{r^* - f(q)\},$$

where $R_{EE} > 0$ is a designer-defined positive-definite constant matrix. There are many ways to handle (4), such as the Newton–Raphson and various metaheuristics-based strategies. However, as discussed in Section I-B, the traditional methods for (4) involve some difficulties, which degrade their usability. In particular, note that merely solving (4) does not consider the joint motion cost: it may result in excessive joint motion, which is practically undesirable. This is because although (3) (and hence (4)) generally has multiple solutions, all of which realize the desired position and orientation of the end-effector, not every solution yields the same posture of the manipulator itself. Users must choose the appropriate solution for efficient robot operation.

*B. Consideration of Motion Cost*

As discussed in Remark 4, simply solving (4) has a significant drawback from the viewpoint of motion cost. This study proposes a novel formulation of the IK computation considering the joint motion efficiency. Specifically, we propose the formulation of the IK problem as follows:

$$\begin{aligned} &\underset{q}{\text{minimize}} && J_{JMC}(q) \\ &\text{subject to} && J_{EE}(q) = 0, \end{aligned} \quad (5)$$

where $J_{JMC}(q) > 0$ is a scalar function to evaluate the joint motion cost. The specific choice of $J_{JMC}(q)$ depends on the users' requirements. Note that the solutions for (5) are also the solutions for (3), i.e., the original IK problem, because $J_{EE}(q) = 0$ if and only if $\epsilon(q, r^*) = 0$. Consequently, instead of directly minimizing $J_{EE}(q)$, the proposed method minimizes the joint motion cost under the constraint of achieving the desired position and orientation of the end-effector.

Directly solving (5) is challenging because the constraint $J_{EE}(q) = 0$ is a complicated nonlinear equation, which is the original IK problem itself. Therefore, in this study, constrained problem (5) is converted into an unconstrained optimization problem. Specifically, focusing on the fact that $J_{EE}(q) \geq 0$ for all $q$, $J_{EE}(q)$ is regarded as a penalty term and added to the function $J_{JMC}(q)$ to be minimized. Then, the unconstrained optimization problem for the IK computation considering the joint motion efficiency is defined as follows:

$$\underset{q}{\text{minimize}} \quad J(q) \quad (6)$$

$$J(q) := \widetilde{W}_{JMC} J_{JMC}(q) + \widetilde{W}_{EE} J_{EE}(q) \quad (7)$$

$$\widetilde{W}_{JMC} := \frac{W_{JMC}}{W_{JMC} + W_{EE}} \quad (8)$$

$$\widetilde{W}_{EE} := \frac{W_{EE}}{W_{JMC} + W_{EE}}, \quad (9)$$

where $W_{JMC} \in \mathbb{R}_+ \cup \{0\}$ and $W_{EE} \in \mathbb{R}_+$ are constant weights for evaluating the relative importance of the joint motion efficiency and the accuracy of the position and orientation of the end-effector provided by the IK solution, respectively. Larger $W_{JMC}$ value yields more efficient joint motion; larger $W_{EE}$ realizes more accurate position and orientation of the end-effector. Their specific values are determined by a designer according to the requirements of the system. Note that $W_{EE} > 0$ because, in case of $W_{EE} = 0$, $J_{EE}(q)$ is not reflected in the $J(q)$ and minimizing $J(q)$ does not result in solving the original IK problem (4) at all. The relative importance of the joint motion efficiency and the accuracy of the resulting end-effector position and orientation can be altered by adjusting $W_{JMC}$ and $W_{EE}$.

## III. PROPOSED IK COMPUTATION ALGORITHM

Problem (6) is a nonlinear optimization problem. A wide variety of optimization algorithms (e.g., the GA, the PSO method) may be used for solving (6). However, complicated algorithms should be avoided owing to their implementation burden. This study proposes an IK computation algorithm based on NLSPSA owing to its simplicity.

*A. Simultaneous Perturbation Stochastic Approximation and NLSPSA: an Overview*

This section presents an overview of the simultaneous perturbation stochastic approximation (SPSA) and NLSPSA, which is a variant of SPSA. SPSA is an optimization algorithm classified as the stochastic approximation framework [42]. Here, let $\varphi \in \mathbb{R}^p$ be a design variable, $L(\varphi): \mathbb{R}^p \to \mathbb{R}_+ \cup \{0\}$ be the loss function to be minimized, and $\hat{L}(\varphi)$ be its measured value at $\varphi$. Let $\hat{g}(\varphi)$ be a stochastic approximation of $\partial L / \partial \varphi$ (i.e., the gradient of $L$). The update rule for $\varphi$ in the standard SPSA algorithm is described as follows [43]:

$$\varphi_{k+1} = \varphi_k - a_k \hat{g}_k(\varphi_k) \quad (10)$$

$$\hat{g}_k(\varphi_k) = \frac{\hat{L}(\varphi_k + c_k \delta_k) - \hat{L}(\varphi_k - c_k \delta_k)}{2c_k} \delta_k^{(-1)} \quad (11)$$

$$\delta_k^{(-1)} := \left[ \frac{1}{\delta_{k,1}} \quad \frac{1}{\delta_{k,2}} \quad \cdots \quad \frac{1}{\delta_{k,p}} \right]^T, \quad (12)$$

where the index $k$ denotes the $k$th iteration; $\delta_k \in \mathbb{R}^p$ is a random perturbation vector satisfying certain statistical properties (a Bernoulli $\pm 1$ distribution with a probability of 0.5 for each $\pm 1$ outcome is often employed as $\delta_k$ [44]) and $\delta_{k,i}$ ($i = 1, 2, \cdots p$) is the $i$th element of $\delta_k$; and $a_k$ and $c_k$ are positive gain sequences. Equations (13) and (14) are typical settings for $a_k$ and $c_k$, respectively [42]:

$$a_k = \frac{a}{(A + k)^\alpha} \quad (13)$$

$$c_k = \frac{c}{k^\gamma}, \quad (14)$$

where $a$, $A$, $c$, $\alpha$, and $\gamma$ are positive constants.

The most representative feature of SPSA is its high calculation efficiency. Specifically, (10) and (11) indicate that the number of measurements of the loss function per iteration is only two regardless of the problem dimension. In spite of such a small number of required loss measurements, the convergence rate of SPSA is the same as that of the finite-difference stochastic approximation, which requires $2p$ measurements of the loss function values, under reasonably general conditions [42]. Accordingly, SPSA is a powerful and efficient technique for high-dimensional optimization problems [45]. Owing to its high computation efficiency, simple algorithm structure, and good searching ability, SPSA is used in many engineering fields, such as adaptive control [46], [47], controller tuning [48]–[51], and consensus algorithm [52]. Moreover, a practical guidelines for implementing SPSA has been provided [42].

Recently, a previous study has pointed out that the conventional SPSA algorithm expressed by (10)–(14) has scope for improvement in terms of the calculation stability for practical applications [53]. To improve the calculation stability, a modified SPSA algorithm called NLSPSA has been proposed [53]:

$$\varphi_{k+1} = \varphi_k - sat\{a_k \hat{g}_k(\varphi_k)\}, \qquad (15)$$

where $sat\{x\}: \mathbb{R}^p \rightarrow \mathbb{R}^p$ for $x \in \mathbb{R}^p$ is defined as

$$sat\{x\} := \begin{bmatrix} \text{sgn}(x_1)\min(|x_1|, d) \\ \text{sgn}(x_2)\min(|x_2|, d) \\ \vdots \\ \text{sgn}(x_p)\min(|x_p|, d) \end{bmatrix}, \qquad (16)$$

where $d$ is a positive constant. In NLSPSA, $\delta_k$ is typically set as a Bernoulli $\pm 1$ distribution with a probability of 0.5 for each $\pm 1$ outcome. A previous study has theoretically analyzed the convergence of NLSPSA and has numerically compared it with the traditional SPSA expressed by (10)–(14), demonstrating that NLSPSA outperforms the traditional SPSA in terms of calculation stability [53]. This study employs NLSPSA for IK computation owing to its calculation efficiency and stability.

### B. Proposed IK Computation Scheme

Let $q^0$ and $q^*$ be the current joint configuration and the desired joint configuration, respectively, i.e., $q^*$ is the solution of (6) to bring the position and orientation of the end-effector to $r^*$ while $J_{JMC}$ is minimized. The proposed technique computes $q^*$ numerically. Let $\varphi_k$ be the estimation of $q^*$ at the $k$th iteration in the proposed algorithm and $N\_max \in \mathbb{N}$ be the maximum number of iterations. The pseudo-code for the proposed computation scheme is shown in Algorithm 1. In Algorithm 1, $a_k$ and $c_k$ are given by (13) and (14), respectively, and $\delta_k$ is a Bernoulli $\pm 1$ distribution with a probability of 0.5 for each $\pm 1$ outcome. If the joint motion range is specified, i.e., $q^* \in [q_{min} \quad q_{max}]$, it can be easily imposed by inserting $\varphi_{k+1} = \min(\varphi_{k+1}, q_{max})$ and $\varphi_{k+1} = \max(\varphi_{k+1}, q_{min})$ after line 8 in Algorithm 1.

*Remark 5*: SPSA (and hence NLSPSA) is a stochastic realization of the gradient-based optimization scheme. However, this technique requires only the loss function values; explicit computation of the gradient of the loss function is not necessary. Thus, the proposed technique, Algorithm 1, does not require explicit calculations of the gradient of $J(q)$. Furthermore, the computation of the gradient of the FK function $f$ is not necessary depending on the selection of $J_{JMC}$.

Therefore, the proposed technique is simpler and easier than traditional gradient-based IK computation methods. Note that the required number of loss function computation is only two at each iteration in the proposed algorithm, which is independent of the degree of freedom of the manipulator, showing its good computational efficiency. In addition, the stochastic approximation of the gradient is expected to improve the computational stability. The proposed algorithm proceeds the iteration based on the stochastically approximated gradient of the loss function. This approximated gradient is, more or less, different from the true derivative in each iteration. We can expect that such an instantaneous approximation error helps to avoid the numerical instability due to the singularity. In fact, the proposed algorithm successfully handles the IK computation in or near a singular configuration, as demonstrated in the numerical examples (see Examples #1.7, #1.8, and #2.3 in Section IV).

*Remark 6*: In the context of IK computations, several studies have focused on solving the IK problem for serial manipulators via a simultaneous perturbation algorithm owing to its low computation cost and good searching ability [54], [55]. However, these studies have used the algorithm not for directly computing the IK but for constructing neural networks that approximate the IK function of the manipulator. Therefore, unlike the proposed strategy, they have significant limitations from the viewpoint of versatility, as discussed in Section I-B.

| **Algorithm 1:** The Proposed IK Algorithm |
|---|
| 1    Input $r^*$ and $q^0$. |
| 2    Initialize $a_k$, $c_k$, and $\delta_k$. |
| 3    $\varphi_1 = q^0$ |
| 4    **for** $k = 1: N\_max$ |
| 5          Generate $a_k$, $c_k$, and $\delta_k$. |
| 6          Compute $J(\varphi_k + c_k\delta_k)$ and $J(\varphi_k - c_k\delta_k)$. |
| 7          $\hat{g}_k(\varphi_k) = \dfrac{J(\varphi_k + c_k\delta_k) - J(\varphi_k - c_k\delta_k)}{2c_k}\delta_k^{(-1)}$ |
| 8          Update $\varphi_k$ by Eq. (15). |
| 9    **end** |
| 10    Return $\varphi_{N\_max}$ as the computation result for $q^*$. |

## IV. VALIDATION OF THE PROPOSED TECHNIQUE

The effectiveness of the proposed approach (Algorithm 1) is demonstrated by applying it to IK computations for planar redundant manipulators. The position and orientation of the end-effector are denoted by $r = [x \quad y \quad \theta]^T$. Hereafter, $J_{JMC}$ is set as

$$J_{JMC}(q) = (q - q^0)^T Q_{JMC}(q - q^0), \qquad (17)$$

where $Q_{JMC}$ is a positive-definite constant matrix having a compatible dimension. Note that the setting of $J_{JMC}$ is not limited to (17) in Algorithm 1 as the proposed algorithm does not depend on a specific form of $J_{JMC}$. The parameter values for NLSPSA employed in this study are $A = 10$, $a = 3.0 \times 10^3$, $c = 0.1$, $\alpha = 0.602$, $\gamma = 0.101$, and $d = 0.03$.

The maximum number of iterations $N\_max$ is 25000. Further, $W_{JMC}$ and $W_{EE}$ are set to 1 and 50, respectively, and $R_{EE} = (diag\{1 \quad 1 \quad 5(2\pi/360)^2\})/7$. The limitation of joint motion range is not imposed.

*Remark 7:* The joint motion cost $J_{JMC}$ in (17) can control each joint motion amount by appropriately setting $Q_{JMC}$. For example, if $Q_{JMC}$ is set as a diagonal matrix, the motion of a specific joint can be reduced by setting the corresponding element of $Q_{JMC}$ to be larger than the others. This mechanism is useful for evaluating and minimizing the energy consumption of the manipulator because, in general, the required energy for driving a joint is not the same depending on the manipulator geometry and task. Moreover, the specific type of $J_{JMC}$ is not limited to (17); other types of $J_{JMC}$ may be preferred in some situations.

### A. 8-DOF Manipulator

The proposed scheme is applied to an 8-DOF manipulator. All its joints are revolute joints, and all its links have unit length. Note that this manipulator has redundant degrees of freedom: its IK problem has infinite solutions in general. Therefore, one must determine an appropriate solution from the set of solutions, which complicates the IK problem in practice.

Let $q = [q_1 \quad q_2 \quad \cdots \quad q_8]^T$ be the joint angle vector ($q_i$ [deg.] corresponds to the angle of the $i$th joint numbered from the base similar to Fig. 1). The FK function $f_{8DOF}(q)$ of this manipulator can be written as

$$r = \begin{bmatrix} x \\ y \\ \theta \end{bmatrix} = f_{8DOF}(q) = \begin{bmatrix} \sum_{p=1}^{8} \cos\left\{\sum_{i=1}^{p} q_i\right\} \\ \sum_{p=1}^{8} \sin\left\{\sum_{i=1}^{p} q_i\right\} \\ mod\left\{\left(\sum_{p=1}^{8} q_i\right), 360\right\} \end{bmatrix}, \quad (18)$$

where the modulo operation $mod$ is defined as

$$mod\{\alpha, \beta\} := \alpha - \beta \left\lfloor \frac{\alpha}{\beta} \right\rfloor, \quad (19)$$

where $\lfloor \cdot \rfloor$ is the floor function and $\alpha \in \mathbb{R}, \beta \in \mathbb{Z}$.

Table 1 summarizes the validation conditions. Examples #1.1–#1.5 stand for basic movements, which are frequently observed during the operation of the manipulator on site: position alternation without changing the orientation (Examples #1.1–#1.3), orientation alternation without position change (Examples #1.4), and simultaneous alternation of position and orientation (Example #1.5). These examples verify the effectiveness of the proposed IK computation scheme for usual manipulator operations. The aim of Example #1.6 is to show the effectiveness of the consideration of motion cost, which is the main topic of this study. Specifically, the only difference between Examples #1.5 and

#1.6 is the setting of $Q_{JMC}$: the motion cost of each joint is equally evaluated in Example #1.5, whereas driving the joint $q_1$ (the joint that is the closest to the base) imposes higher cost than others in Example #1.6. Therefore, comparing the computational results of Examples #1.5 and #1.6 confirms the validity of the proposed method considering motion cost. Note that the manipulator takes a singular posture in Examples #1.7 and #1.8, which is a challenging situation in numerical IK computation. Hence, Examples #1.7 and #1.8 clearly demonstrate the computation stability of the proposed scheme.

The computation results obtained using the proposed strategy are shown in Figs. 2–5, where the green dot indicates the $(x, y)$ position of $r^*$, the blue line represents the initial posture of the manipulator, and the magenta line represents the resulting posture of the manipulator via the computation result provided by the proposed algorithm. The details of the computation results are summarized in Table 2. Figs. 2–5 and Table 2 clearly show that the proposed computation scheme successfully decreases the loss function value and provides the joint variable values for realizing the desired position and orientation of the end-effector. In Fig. 4, comparison of the motion of the joint that is closest to the base of Example #1.5 with that of Example #1.6 clearly shows the effectiveness of the consideration of the joint motion cost, which is the main topic of this study. That is, owing to $J_{JMC}(q)$, the economic joint motion is realized according to the evaluation of each joint motion cost. Moreover, in Examples #1.7 and #1.8, the computations do not fail even though the manipulator is initially in the singularity state, indicating the good stability of the proposed algorithm.

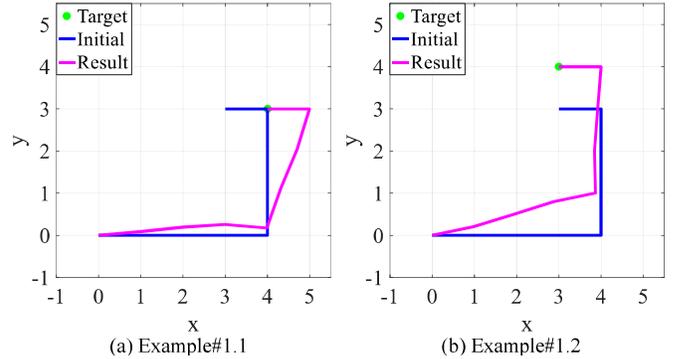

**Fig. 2.** Result of Examples #1.1 and #1.2.

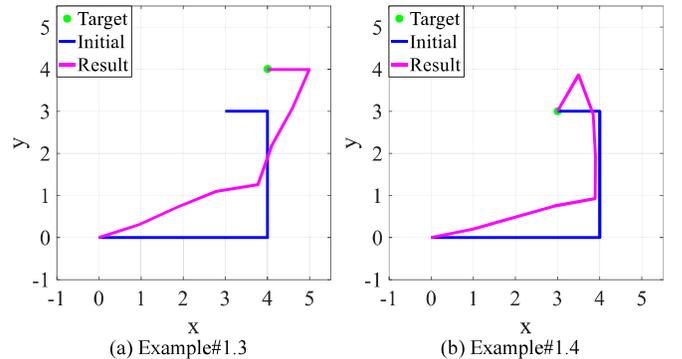

**Fig. 3.** Result of Examples #1.3 and #1.4.

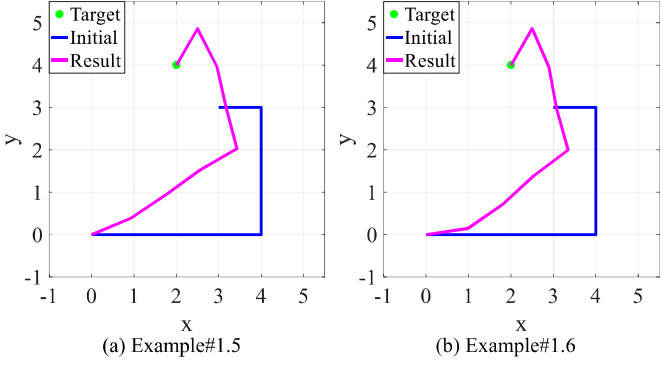

(a) Example#1.5     (b) Example#1.6

**Fig. 4.** Result of Examples #1.5 and #1.6.

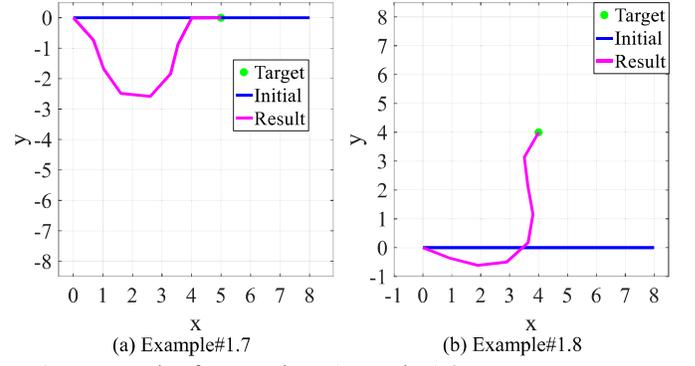

(a) Example#1.7     (b) Example#1.8

**Fig. 5.** Result of Examples #1.7 and #1.8.

TABLE 1
VALIDATION SETTING FOR EXAMPLES #1.1–#1.8.

| Example | Initial joint configuration $q^0$ [deg.] | Initial position and orientation of end-effector $f_{8DOF}(q^0) = [x^0 \quad y^0 \quad \theta^0]^T$ | Desired position and orientation of end-effector $r^* = [x^* \quad y^* \quad \theta^*]^T$ | $Q_{JMC}$ |
|---|---|---|---|---|
| #1.1 | $[0 \quad 0 \quad 0 \quad 0 \quad 90 \quad 0 \quad 0 \quad 90]^T$ | $[3 \quad 3 \quad 180°]^T$ | $[4 \quad 3 \quad 180°]^T$ | $\frac{1}{8}\left(\frac{2\pi}{360}\right)^2 diag\{1 \quad 1 \quad 1 \quad 1 \quad 1 \quad 1 \quad 1 \quad 1\}$ |
| #1.2 | $[0 \quad 0 \quad 0 \quad 0 \quad 90 \quad 0 \quad 0 \quad 90]^T$ | $[3 \quad 3 \quad 180°]^T$ | $[3 \quad 4 \quad 180°]^T$ | $\frac{1}{8}\left(\frac{2\pi}{360}\right)^2 diag\{1 \quad 1 \quad 1 \quad 1 \quad 1 \quad 1 \quad 1 \quad 1\}$ |
| #1.3 | $[0 \quad 0 \quad 0 \quad 0 \quad 90 \quad 0 \quad 0 \quad 90]^T$ | $[3 \quad 3 \quad 180°]^T$ | $[4 \quad 4 \quad 180°]^T$ | $\frac{1}{8}\left(\frac{2\pi}{360}\right)^2 diag\{1 \quad 1 \quad 1 \quad 1 \quad 1 \quad 1 \quad 1 \quad 1\}$ |
| #1.4 | $[0 \quad 0 \quad 0 \quad 0 \quad 90 \quad 0 \quad 0 \quad 90]^T$ | $[3 \quad 3 \quad 180°]^T$ | $[3 \quad 3 \quad 240°]^T$ | $\frac{1}{8}\left(\frac{2\pi}{360}\right)^2 diag\{1 \quad 1 \quad 1 \quad 1 \quad 1 \quad 1 \quad 1 \quad 1\}$ |
| #1.5 | $[0 \quad 0 \quad 0 \quad 0 \quad 90 \quad 0 \quad 0 \quad 90]^T$ | $[3 \quad 3 \quad 180°]^T$ | $[2 \quad 4 \quad 240°]^T$ | $\frac{1}{8}\left(\frac{2\pi}{360}\right)^2 diag\{1 \quad 1 \quad 1 \quad 1 \quad 1 \quad 1 \quad 1 \quad 1\}$ |
| #1.6 | $[0 \quad 0 \quad 0 \quad 0 \quad 90 \quad 0 \quad 0 \quad 90]^T$ | $[3 \quad 3 \quad 180°]^T$ | $[2 \quad 4 \quad 240°]^T$ | $\frac{1}{57}\left(\frac{2\pi}{360}\right)^2 diag\{50 \quad 1 \quad 1 \quad 1 \quad 1 \quad 1 \quad 1 \quad 1\}$ |
| #1.7 | $[0 \quad 0 \quad 0 \quad 0 \quad 0 \quad 0 \quad 0 \quad 0]^T$ | $[8 \quad 0 \quad 0°]^T$ | $[5 \quad 0 \quad 0°]^T$ | $\frac{1}{8}\left(\frac{2\pi}{360}\right)^2 diag\{1 \quad 1 \quad 1 \quad 1 \quad 1 \quad 1 \quad 1 \quad 1\}$ |
| #1.8 | $[0 \quad 0 \quad 0 \quad 0 \quad 0 \quad 0 \quad 0 \quad 0]^T$ | $[8 \quad 0 \quad 0°]^T$ | $[4 \quad 4 \quad 60°]^T$ | $\frac{1}{8}\left(\frac{2\pi}{360}\right)^2 diag\{1 \quad 1 \quad 1 \quad 1 \quad 1 \quad 1 \quad 1 \quad 1\}$ |

TABLE 2
RESULTS FOR EXAMPLES #1.1–#1.8 DUE TO THE PROPOSED TECHNIQUE.

| Example | Calculation result $\varphi_{N\_max}$ [deg.] | Resulting position and orientation of end-effector provided by the calculation result $f_{8DOF}(\varphi_{N\_max}) = [x \quad y \quad \theta]^T$ | Initial loss function value $J(\varphi_1)$ | Final loss function value $J(\varphi_{N\_max})$ |
|---|---|---|---|---|
| #1.1 | $[5.0845 \quad 0.9802 \quad -2.2995 \quad -8.6481$ $75.5098 \quad -3.1621 \quad 5.2970 \quad 107.1592]^T$ | $[3.9960 \quad 2.9983 \quad 179.9210°]^T$ | 0.1401 | $4.8879 \times 10^{-4}$ |
| #1.2 | $[12.0536 \quad 4.9613 \quad 0.3230 \quad -5.5771$ $79.5778 \quad -5.6460 \quad -0.2131 \quad 94.4796]^T$ | $[2.9983 \quad 3.9978 \quad 179.9591°]^T$ | 0.1401 | $2.7151 \times 10^{-4}$ |
| #1.3 | $[17.8475 \quad 6.7628 \quad -2.6768 \quad -12.7819$ $61.5466 \quad -9.7258 \quad 5.4850 \quad 113.3832]^T$ | $[3.9911 \quad 3.9932 \quad 179.8407°]^T$ | 0.2801 | $1.5279 \times 10^{-3}$ |
| #1.4 | $[11.3359 \quad 4.6545 \quad -0.1554 \quad -5.3907$ $78.9524 \quad 3.6319 \quad 17.3524 \quad 129.5170]^T$ | $[2.9952 \quad 2.9970 \quad 239.8979°]^T$ | 0.7679 | $1.6323 \times 10^{-3}$ |
| #1.5 | $[22.6410 \quad 11.0881 \quad 2.3066 \quad -6.1528$ $74.8829 \quad -1.8553 \quad 14.0732 \quad 122.9056]^T$ | $[1.9966 \quad 3.9945 \quad 239.8893°]^T$ | 1.0481 | $1.6447 \times 10^{-3}$ |
| #1.6 | $[8.3777 \quad 26.4987 \quad 7.6871 \quad -5.5986$ $68.3646 \quad -4.7470 \quad 12.8459 \quad 126.5138]^T$ | $[1.9988 \quad 3.9947 \quad 239.9422°]^T$ | 1.0481 | $6.6408 \times 10^{-4}$ |
| #1.7 | $[-47.2853 \quad -22.6027 \quad 15.9377 \quad 48.0562$ $52.9799 \quad 28.3487 \quad -13.8128 \quad -61.3942]^T$ | $[5.0131 \quad -9.3435 \times 10^{-4} \quad 0.2273°]^T$ | 1.2605 | $9.6520 \times 10^{-3}$ |
| #1.8 | $[-21.2191 \quad 6.7520 \quad 21.1489 \quad 35.4891$ $37.9137 \quad 20.0604 \quad -3.1548 \quad -36.8400]^T$ | $[4.0069 \quad 4.0053 \quad 60.1502°]^T$ | 5.2497 | $4.0543 \times 10^{-3}$ |

## B. 20-DOF Manipulator

To further emphasize the effectiveness of the proposed approach, this study tackles the IK problem of a 20-DOF manipulator using the proposed algorithm. As with the 8-DOF manipulator, all its joints are revolute joints, and all its links have unit length. Note that this 20-DOF manipulator has many more redundant degrees of freedom than the 8-DOF manipulator examined in Section IV-A, which significantly increases the difficulty in the selection of a suitable IK solution. Further, note that nearly no modification is required for the proposed algorithm for conversion from the 8-DOF example to the 20-DOF example, which indicates its high versatility. Therefore, the 20-DOF case in this section together with the 8-DOF case in the previous section clearly demonstrate the computational ability and usability of the proposed approach.

Similar to the 8-DOF case, the FK mapping $f_{20DOF}(q)$ of this manipulator can be expressed by (20). The validation conditions are listed in Table 3, where $I_{20}$ is a $20 \times 20$ identity matrix. Note that in Example #2.2, the desired position and orientation of the end-effector is achievable only if the manipulator takes a nearly singular posture, and in Example #2.3, the initial posture of the manipulator is singular. These situations generally make the IK computation much more challenging.

$$r = \begin{bmatrix} x \\ y \\ \theta \end{bmatrix} = f_{20DOF}(q) = \begin{bmatrix} \sum_{p=1}^{20} \cos\left\{\sum_{i=1}^{p} q_i\right\} \\ \sum_{p=1}^{20} \sin\left\{\sum_{i=1}^{p} q_i\right\} \\ mod\left\{\left(\sum_{p=1}^{20} q_i\right), 360\right\} \end{bmatrix}, \quad (20)$$

The computation results are shown in Fig. 6, where the meanings of the colors are the same as those in Figs. 2–5. The details of the computation results are summarized in Table 4. These results clearly show that the proposed method is effective for the IK computation of not only the 8-DOF manipulator but also the 20-DOF manipulator.

Finally, we try to address the optimization problem (6) via PSO so as to demonstrate the computational efficiency of the NLSPSA-based approach (Algorithm 1). Specifically, for Examples #2.1–#2.3, the optimization problem (6) is solved via PSO implemented in the MATLAB Global Optimization Toolbox. PSO is a typical population-based metaheuristics algorithm and is often employed to solve challenging engineering optimization problems including the IK problem. Note that the comparison between NLSPSA and PSO is fair from the viewpoint of the required prior knowledge for the problem: both algorithms require only the evaluation of loss function value and do not need an analytical derivative of loss function. Furthermore, to make the comparison fair, the budget for loss function evaluation is $50000 (= 2 \times N\_max)$, which is the same as that for the proposed scheme. The population size is 100, and the initial population is generated as $q^0 \pm 20 \times rand(20,1)$, where $rand(20,1) \in \mathbb{R}^{20}$ denotes a random vector (each element is uniformly distributed in interval $(0,1)$). As a result, the minimal loss function values found through the optimization process are $1.6526 \times 10^{-3}$ (Example #2.1), $2.8847 \times 10^{-3}$ (Example #2.2), and $2.2372 \times 10^{-3}$ (Example #2.3). These results and the results in Table 4 clearly show that NLSPSA outperforms PSO in the proposed IK problem (6). Consequently, the comparative study supports the high computational efficiency of Algorithm 1.

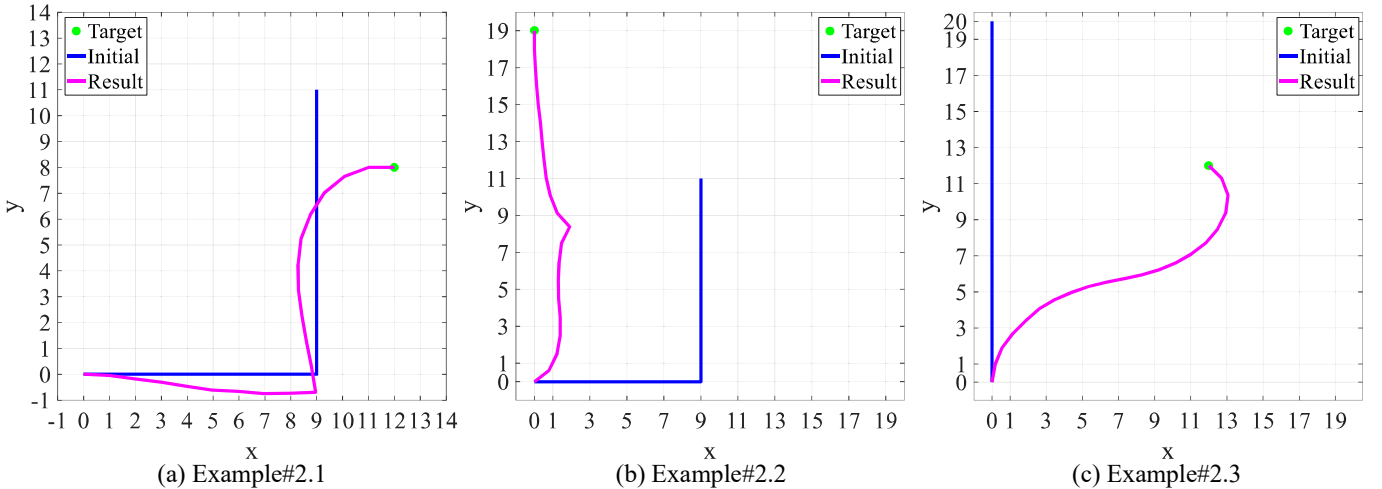

(a) Example#2.1   (b) Example#2.2   (c) Example#2.3

**Fig. 6.** Result of Examples #2.1–#2.3.

TABLE 3
VALIDATION SETTING FOR EXAMPLES #2.1–#2.3.

| Example | Initial joint configuration $q^0$ [deg.] | Initial position and orientation of end-effector $f_{20DOF}(q^0)=[x^0 \ \ y^0 \ \ \theta^0]^T$ | Desired position and orientation of end-effector $r^*=[x^* \ \ y^* \ \ \theta^*]^T$ | $Q_{JMC}$ |
|---|---|---|---|---|
| #2.1 | $[0 \ \ 0 \ \ 0 \ \ 0 \ \ 0 \ \ 0 \ \ 0 \ \ 0 \ \ 0 \ \ 90$ <br> $0 \ \ 0 \ \ 0 \ \ 0 \ \ 0 \ \ 0 \ \ 0 \ \ 0 \ \ 0 \ \ 0]^T$ | $[9 \ \ 11 \ \ 90°]^T$ | $[12 \ \ 8 \ \ 0°]^T$ | $\frac{1}{20}\left(\frac{2\pi}{360}\right)^2 I_{20}$ |
| #2.2 | $[0 \ \ 0 \ \ 0 \ \ 0 \ \ 0 \ \ 0 \ \ 0 \ \ 0 \ \ 0 \ \ 90$ <br> $0 \ \ 0 \ \ 0 \ \ 0 \ \ 0 \ \ 0 \ \ 0 \ \ 0 \ \ 0 \ \ 0]^T$ | $[9 \ \ 11 \ \ 90°]^T$ | $[0 \ \ 19 \ \ 90°]^T$ | $\frac{1}{20}\left(\frac{2\pi}{360}\right)^2 I_{20}$ |
| #2.3 | $[90 \ \ 0 \ \ 0 \ \ 0 \ \ 0 \ \ 0 \ \ 0 \ \ 0 \ \ 0 \ \ 0$ <br> $0 \ \ 0 \ \ 0 \ \ 0 \ \ 0 \ \ 0 \ \ 0 \ \ 0 \ \ 0 \ \ 0]^T$ | $[0 \ \ 20 \ \ 90°]^T$ | $[12 \ \ 12 \ \ 135°]^T$ | $\frac{1}{20}\left(\frac{2\pi}{360}\right)^2 I_{20}$ |

TABLE 4
RESULTS FOR EXAMPLES #2.1–#2.3 DUE TO THE PROPOSED TECHNIQUE.

| Example | Calculation result $\varphi_{N\_max}$ [deg.] | Resulting position and orientation of end-effector provided by the calculation result $f_{20DOF}(\varphi_{N\_max})=[x \ \ y \ \ \theta]^T$ | Initial loss function value $J(\varphi_1)$ | Final loss function value $J(\varphi_{N\_max})$ |
|---|---|---|---|---|
| #2.1 | $[-2.7917 \ \ -4.7071 \ \ 0.3216 \ \ -2.1181$ <br> $0.8099 \ \ 5.7666 \ \ -1.9838 \ \ 5.3438$ <br> $1.2744 \ \ 96.6482 \ \ 3.3685 \ \ -2.2174$ <br> $-2.2785 \ \ -6.1927 \ \ -7.8593 \ \ -14.7877$ <br> $-10.8078 \ \ -18.5754 \ \ -18.7799 \ \ -20.1424]^T$ | $[12.0027 \ \ 8.0031 \ \ 0.2913°]^T$ | $4.2489$ | $5.3035 \times 10^{-4}$ |
| #2.2 | $[38.0090 \ \ 26.1704 \ \ 15.8720 \ \ 9.9746$ <br> $4.6203 \ \ -3.3546 \ \ -3.7743 \ \ -4.8520$ <br> $-18.9813 \ \ 68.3634 \ \ -20.3324 \ \ -8.2872$ <br> $-6.4977 \ \ -1.3128 \ \ -0.2449 \ \ 0.9352$ <br> $-0.9134 \ \ -1.3930 \ \ -1.2213 \ \ -2.7694]^T$ | $[6.9225 \times 10^{-5} \ \ 18.9983 \ \ 90.0105°]^T$ | $20.3081$ | $1.1707 \times 10^{-3}$ |
| #2.3 | $[79.8296 \ \ -12.5086 \ \ -14.2108 \ \ -8.2644$ <br> $-3.4241 \ \ -10.9131 \ \ -7.4559 \ \ -4.1181$ <br> $-5.2305 \ \ -2.5847 \ \ 1.0087 \ \ 4.6744$ <br> $5.7501 \ \ 6.9117 \ \ 7.9734 \ \ 12.2353$ <br> $13.3835 \ \ 19.5450 \ \ 27.7220 \ \ 24.6442]^T$ | $[11.9998 \ \ 12.0003 \ \ 134.9677°]^T$ | $29.5636$ | $9.0259 \times 10^{-4}$ |

## V. DISCUSSION

The validation results in Section IV clearly show that the proposed approach can compute the IK of serial manipulators with redundancy. Note that the results confirm that the proposed method is applicable to both 8-DOF and 20-DOF manipulator with nearly no modifications, and it can handle various movements. Accordingly, the proposed scheme is flexible and versatile, and it is especially useful when dealing with multiple manipulators with various geometries. This is because the computation strategy requires neither analytical derivation of the gradient of FK mapping for each manipulator nor training of neural networks. This feature shows that this computation strategy has a significant advantage over traditional gradient-based methods and learning-based approaches using neural networks mentioned in Section I-B. Moreover, the proposed algorithm (Algorithm 1) requires only two computations of the loss function value at each iteration regardless of the degrees of freedom of the manipulator. Therefore, this algorithm exhibits high computational efficiency especially for highly redundant manipulators.

Inspired by the IK problem in the field of biological cybernetics, the proposed method explicitly considers the joint motion cost of the manipulator via $J_{JMC}(q)$. The effectiveness of introducing $J_{JMC}(q)$ is clearly verified in the comparison of Examples #1.5 and #1.6 shown in Fig. 4. The main difference between Examples #1.5 and #1.6 is the relative magnitude of each diagonal component of $Q_{JMC}$. Specifically, the motion cost of each joint is equally evaluated in Example #1.5. Meanwhile, the motion of $q_1$ (the joint that is the closest to the base) is given a larger weight than the others so as to restrict its motion in Example #1.6. In fact, the motion of $q_1$ in Example #1.6 is significantly smaller than that in Example #1.5, clearly demonstrating the effectiveness of $J_{JMC}(q)$ for efficient manipulator operation.

SPSA (and hence NLSPSA) can handle various types of loss functions because it requires only measured values of the loss function, even though it is a stochastic realization of a gradient-based optimization. From this viewpoint, the proposed technique can be regarded as an IK computation strategy that bridges the gap between gradient-based and the metaheuristics-based methods. Therefore, various measures of the joint motion cost can be employed for $J_{JMC}(q)$ in (7) depending on the mechanism, geometry, etc., of the manipulator; it is not restricted to (17). Moreover, although the self-collision and obstacle avoidance are not considered in this study, these additional constraints will be handled by including an appropriate term in $J_{JMC}(q)$ owing to the simplicity and wide applicability of the SPSA algorithm. In the future, various $J_{JMC}(q)$ will be examined for IK problems with challenging constraints such as self-collision.

In Examples #1.7, #1.8, #2.2, and #2.3, the initial or final posture of the manipulator is singular or nearly singular.

Nevertheless, the proposed approach successfully computes the appropriate solution of the IK problem. Therefore, the proposed algorithm is numerically stable under singularity. This practically useful characteristic originates from two features of the algorithm: the stochastic property of NLSPSA and the saturation mechanism (16). In particular, NLSPSA used in this study stochastically approximates the gradient of the loss function, as shown in (11). Therefore, an approximation error between (11) and the true gradient exists especially in the early iterations; this random error leads to lower sensitivity to numerical instability and facilitates the exit from the unstable region. Note that SPSA can become a global optimizer owing to this randomness [56]; hence it has a significant advantage over traditional gradient methods. Moreover, NLSPSA has the saturation mechanism (16) so as to avoid excessive parameter update, thereby enhancing the calculation stability under ill conditions.

One main contribution of this study is to propose the novel formulation of the IK computation problem: computation of practical solution from the viewpoint of joint motion efficiency. To solve the proposed IK problem, the simple computation scheme is developed based on NLSPSA. Reinforcement learning (RL) techniques [1], [57] such as adaptive dynamic programming (ADP) [58], [59] may be another suitable strategy to tackle the proposed IK problem, since it can overcome the difficulties in traditional learning-based approaches discussed in Section I-B, owing to its flexibility. In fact, several studies have reported the effectiveness of the RL approach for robot systems (e.g., [60], [61]), and a new learning framework has been proposed [62]. In the future, a new computation scheme for the proposed IK problem will be developed based on ADP, and it will be compared with the NLSPSA-based scheme proposed in this study. Although several studies have examined the IK computation via RL [63], no studies have explicitly considered the practical usefulness of the solution, to the best of our knowledge.

Although the proposed method is usually efficient, there is scope for enhancement of the convergence when each design variable has significantly different magnitude from one another. It is because the desired joint configuration does not always have the similar magnitude to each other in real-world industrial scenarios. In the proposed algorithm, the magnitudes of the design variables updated in one iteration are the same as each other, since each element of $\delta_k$ (the simultaneous perturbation vector) is 1 or $-1$ (see (11) and (15)). This feature is disadvantageous for fast convergence when each design variable has very different magnitude from one another. This is a typical challenge for the first-order SPSA algorithm. Recently, several studies have proposed various adaptive parameter update techniques, i.e., adaptive SPSA, so as to address this issue [64], [65]. Adaptive SPSA is the stochastic analogue of the Newton–Raphson algorithm. Note, however, that the adaptive SPSA algorithm is sensitive to the initial and numerical conditions as the conventional Newton–Raphson algorithm and requires more hyperparameters. Therefore, naive introduction of adaptive SPSA to the proposed IK computation may rather decrease the practical usability of the algorithm due to the unstable computation. In the future, we will examine the IK algorithm, which can balance the fast convergence and stable computation, based on adaptive SPSA.

Regarding the inverse calculation problems in biological cybernetics, the proposed computation may be extended to trajectory computation by considering the actuator dynamics. For example, one study has proposed a hypothesis for human arm trajectory generation: the human arm trajectory is generated so as to minimize the required metabolic energy for motion [39], and it is verified on the basis of a mathematical model of metabolic energy. In the future, for more efficient operation of manipulators, the inverse calculation will be further improved by considering the actuator dynamics.

## VI. CONCLUSION AND FUTURE WORK

For efficient operation of a manipulator, this study proposed a simple and efficient IK computation strategy considering the joint motion cost. In the proposed approach, the IK problem is defined as a minimization problem of the joint motion cost, where the end-effector position and orientation are regarded as a constraint. Then, the minimization problem is solved using the NLSPSA algorithm. The effectiveness of the proposed technique was verified by applying it to the IK computation of 8-DOF and 20-DOF manipulators, thereby confirming its flexibility, computation stability, and motion economy of the solution.

In the future, the proposed method will be compared with the existing IK computation methods from the viewpoint of computation efficiency and stability. Finally, the effectiveness of the proposed approach will be evaluated in experiments using a real-world manipulator.

## REFERENCES

[1] Y. Ouyang, L. Dong, and C. Sun, "Critic learning-based control for robotic manipulators with prescribed constraints," *IEEE Trans. Cybern.*, vol. 52, no. 4, pp. 2274–2283, Apr. 2022.

[2] Y. Cong, R. Chen, B. Ma, H. Liu, D. Hou, and C. Yang, "A comprehensive study of 3-D vision-based robot manipulation," *IEEE Trans. Cybern.*, vol. 53, no. 3, pp. 1682–1698, Mar. 2023.

[3] S. Makita, T. Sasaki, and T. Urakawa, "Offline direct teaching for a robotic manipulator in the computational space," *Int. J. Autom. Technol.*, vol. 15, no. 2, pp. 197–205, Mar. 2021.

[4] R. Drath and A. Horch, "Industrie 4.0: Hit or hype? [Industry Forum]," *IEEE Ind. Electron. Mag.*, vol. 8, no. 2, pp. 56–58, Jun. 2014.

[5] I. Ahmed, G. Jeon, and F. Piccialli, "From artificial intelligence to explainable artificial intelligence in industry 4.0: A survey on what, how, and where," *IEEE Trans. Ind. Informatics*, vol. 18, no. 8, pp. 5031–5042, Aug. 2022.

[6] N. Ghodsian *et al.*, "Toward designing an integration architecture for a mobile manipulator in production systems: Industry 4.0," *Procedia CIRP*, vol. 109, pp. 443–448, 2022.

[7] Z. Guo, Y. Zhang, X. Zhao, and X. Song, "CPS-based self-adaptive collaborative control for smart production-logistics systems," *IEEE Trans. Cybern.*, vol. 51, no. 1, pp. 188–198, Jan. 2021.

[8] J. Wan, X. Li, H.-N. Dai, A. Kusiak, M. Martinez-Garcia, and D. Li, "Artificial-intelligence-driven customized manufacturing factory: key technologies, applications, and challenges," *Proc. IEEE*, vol. 109, no. 4, pp. 377–398, Apr. 2021.

[9] I. Kovalenko, E. C. Balta, D. M. Tilbury, and K. Barton, "Cooperative product agents to improve manufacturing system flexibility: A model-based decision framework," *IEEE Trans. Autom. Sci. Eng.*, vol. 20, no. 1, pp. 440–457, Jan. 2023.

[10] A. Rikalovic, N. Suzic, B. Bajic, and V. Piuri, "Industry 4.0 implementation challenges and opportunities: A technological perspective," *IEEE Syst. J.*, vol. 16, no. 2, pp. 2797–2810, Jun. 2022.

[11] L. Jin and Y. Zhang, "G2-type SRMPC scheme for synchronous manipulation of two redundant robot arms," *IEEE Trans. Cybern.*, vol. 45, no. 2, pp. 153–164, Feb. 2015.

[12] C.-A. Cheng, H.-P. Huang, H.-K. Hsu, W.-Z. Lai, and C.-C. Cheng, "Learning the inverse dynamics of robotic manipulators in structured reproducing kernel Hilbert space," *IEEE Trans. Cybern.*, vol. 46, no. 7, pp. 1691–1703, Jul. 2016.

[13] J. Xu, K. Song, Y. He, Z. Dong, and Y. Yan, "Inverse kinematics for 6-DOF serial manipulators with offset or reduced wrists via a hierarchical iterative algorithm," *IEEE Access*, vol. 6, pp. 52899–52910, 2018.

[14] C. Lauretti, T. Grasso, E. de Marchi, S. Grazioso, and G. di Gironimo, "A geometric approach to inverse kinematics of hyper-redundant manipulators for tokamaks maintenance," *Mech. Mach. Theory*, vol. 176, p. 104967, Oct. 2022.

[15] Y.-J. Chen, M.-Y. Ju, and K.-S. Hwang, "A virtual torque-based approach to kinematic control of redundant manipulators," *IEEE Trans. Ind. Electron.*, vol. 64, no. 2, pp. 1728–1736, Feb. 2017.

[16] M. Faroni, M. Beschi, and N. Pedrocchi, "Inverse kinematics of redundant manipulators with dynamic bounds on joint movements," *IEEE Robot. Autom. Lett.*, vol. 5, no. 4, pp. 6435–6442, 2020.

[17] P. Trutman, S. M. El Din, D. Henrion, and T. Pajdla, "Globally optimal solution to inverse kinematics of 7DOF serial manipulator," *IEEE Robot. Autom. Lett.*, vol. 7, no. 3, pp. 6012–6019, Jul. 2022.

[18] A. Reiter, A. Muller, and H. Gattringer, "On higher order inverse kinematics methods in time-optimal trajectory planning for kinematically redundant manipulators," *IEEE Trans. Ind. Informatics*, vol. 14, no. 4, pp. 1681–1690, Apr. 2018.

[19] S. Starke, N. Hendrich, and J. Zhang, "Memetic evolution for generic full-body inverse kinematics in robotics and animation," *IEEE Trans. Evol. Comput.*, vol. 23, no. 3, pp. 406–420, Jun. 2019.

[20] I. Zaplana, H. Hadfield, and J. Lasenby, "Closed-form solutions for the inverse kinematics of serial robots using conformal geometric algebra," *Mech. Mach. Theory*, vol. 173, p. 104835, Jul. 2022.

[21] Y. Tong, J. Liu, Y. Liu, and Y. Yuan, "Analytical inverse kinematic computation for 7-DOF redundant sliding manipulators," *Mech. Mach. Theory*, vol. 155, p. 104006, Jan. 2021.

[22] F. Maric, M. Giamou, A. W. Hall, S. Khoubyarian, I. Petrovic, and J. Kelly, "Riemannian optimization for distance-geometric inverse kinematics," *IEEE Trans. Robot.*, vol. 38, no. 3, pp. 1703–1722, Jun. 2022.

[23] T. Sugihara, "Solvability-unconcerned inverse kinematics by the Levenberg–Marquardt method," *IEEE Trans. Robot.*, vol. 27, no. 5, pp. 984–991, Oct. 2011.

[24] P. Falco and C. Natale, "On the stability of closed-loop inverse kinematics algorithms for redundant robots," *IEEE Trans. Robot.*, vol. 27, no. 4, pp. 780–784, Aug. 2011.

[25] S. Lloyd, R. A. Irani, and M. Ahmadi, "Fast and robust inverse kinematics of serial robots using Halley's method," *IEEE Trans. Robot.*, vol. 38, no. 5, pp. 2768–2780, Oct. 2022.

[26] A.-V. Duka, "Neural network based inverse kinematics solution for trajectory tracking of a robotic arm," *Procedia Technol.*, vol. 12, pp. 20–27, 2014.

[27] A. H. B. Boutchouang, A. Melingui, J. J.-B. M. Ahanda, O. Lakhal, F. B. Motto, and R. Merzouki, "Learning-based approach to inverse kinematics of wheeled mobile continuum manipulators," *IEEE/ASME Trans. Mechatronics*, vol. 27, no. 5, pp. 3451–3462, Oct. 2022.

[28] G. Fang, Y. Tian, Z.-X. Yang, J. M. P. Geraedts, and C. C. L. Wang, "Efficient Jacobian-based inverse kinematics with sim-to-real transfer of soft robots by learning," *IEEE/ASME Trans. Mechatronics*, vol. 27, no. 6, pp. 5296–5306, Dec. 2022.

[29] A. G. Jiokou Kouabon *et al.*, "A learning framework to inverse kinematics of high DOF redundant manipulators," *Mech. Mach. Theory*, vol. 153, p. 103978, Nov. 2020.

[30] J. Xu, W. Liu, and L. Li, "An improved GA-based inverse kinematics solution algorithm for underwater manipulators," in *2022 13th Asian Control Conference (ASCC)*, May 2022, pp. 2091–2096.

[31] S. Tabandeh, W. W. Melek, and C. M. Clark, "An adaptive niching genetic algorithm approach for generating multiple solutions of serial manipulator inverse kinematics with applications to modular robots," *Robotica*, vol. 28, no. 4, pp. 493–507, Jul. 2010.

[32] Z. Zhou, H. Guo, Y. Wang, Z. Zhu, J. Wu, and X. Liu, "Inverse kinematics solution for robotic manipulator based on extreme learning machine and sequential mutation genetic algorithm," *Int. J. Adv. Robot. Syst.*, vol. 15, no. 4, p. 172988141879299, Jul. 2018.

[33] D. Wu, G. Hou, W. Qiu, and B. Xie, "T-IK: An efficient multi-objective evolutionary algorithm for analytical inverse kinematics of redundant manipulator," *IEEE Robot. Autom. Lett.*, vol. 6, no. 4, pp. 8474–8481, Oct. 2021.

[34] Y. Liu, J. Xi, H. Bai, Z. Wang, and L. Sun, "A general robot inverse kinematics solution method based on improved PSO algorithm," *IEEE Access*, vol. 9, pp. 32341–32350, 2021.

[35] R. V. Ram, P. M. Pathak, and S. J. Junco, "Inverse kinematics of mobile manipulator using bidirectional particle swarm optimization by manipulator decoupling," *Mech. Mach. Theory*, vol. 131, pp. 385–405, Jan. 2019.

[36] D. O. Santos, L. Molina, J. G. N. Carvalho, E. A. N. Carvalho, and E. O. Freire, "Modifications of fully resampled PSO in the inverse kinematics of robot manipulators," *IEEE Robot. Autom. Lett.*, vol. 9, no. 2, pp. 1923–1928, Feb. 2024.

[37] J. Chen and X. Li, "Determining human upper limb postures with a developed inverse kinematic method," *Robotica*, vol. 40, no. 11, pp. 4120–4142, Nov. 2022.

[38] J. Zhao, S. Gong, B. Xie, Y. Duan, and Z. Zhang, "Human arm motion prediction in human-robot interaction based on a modified minimum jerk model," *Adv. Robot.*, vol. 35, no. 3–4, pp. 205–218, Feb. 2021.

[39] R. M. Alexander, "A minimum energy cost hypothesis for human arm trajectories," *Biol. Cybern.*, vol. 76, no. 2, pp. 97–105, Feb. 1997.

[40] Z. Zhang, S. Yang, and L. Zheng, "A punishment mechanism-combined recurrent neural network to solve motion-planning problem of redundant robot manipulators," *IEEE Trans. Cybern.*, vol. 53, no. 4, pp. 2177–2185, Apr. 2023.

[41] Z. Mu *et al.*, "Hyper-redundant manipulators for operations in confined space: Typical applications, key technologies, and grand challenges," *IEEE Trans. Aerosp. Electron. Syst.*, vol. 58, no. 6, pp. 4928–4937, Dec. 2022.

[42] J. C. Spall, *Introduction to Stochastic Search and Optimization*. Hoboken, New Jersey: Wiley, 2003.

[43] J. C. Spall, "Multivariate stochastic approximation using a simultaneous perturbation gradient approximation," *IEEE Trans. Automat. Contr.*, vol. 37, no. 3, pp. 332–341, 1992.

[44] L. A. Prashanth, S. Bhatnagar, N. Bhavsar, M. Fu, and S. I. Marcus, "Random directions stochastic approximation with deterministic perturbations," *IEEE Trans. Automat. Contr.*, vol. 65, no. 6, pp. 2450–2465, Jun. 2020.

[45] D. Peng, Y. Chen, and J. C. Spall, "Formal comparison of simultaneous perturbation stochastic approximation and random direction stochastic approximation," in *2023 American Control Conference (ACC)*, 2023, pp. 744–749.

[46] I. Kajiwara, K. Furuya, and S. Ishizuka, "Experimental verification of a real-time tuning method of a model-based controller by perturbations to its poles," *Mech. Syst. Signal Process.*, vol. 107, pp. 396–408, Jul. 2018.

[47] A. Yonezawa, H. Yonezawa, and I. Kajiwara, "Vibration control for various structures with time-varying properties via model-free adaptive controller based on virtual controlled object and SPSA," *Mech. Syst. Signal Process.*, vol. 170, p. 108801, 2022.

[48] M.-B. Rădac, R.-E. Precup, E. M. Petriu, and S. Preitl, "Application of IFT and SPSA to servo system control," *IEEE Trans. Neural Networks*, vol. 22, no. 12, pp. 2363–2375, 2011.

[49] A. Yonezawa, H. Yonezawa, and I. Kajiwara, "Parameter tuning technique for a model-free vibration control system based on a virtual controlled object," *Mech. Syst. Signal Process.*, vol. 165, p. 108313, 2022.

[50] M. A. Ahmad, S. I. Azuma, and T. Sugie, "Performance analysis of model-free PID tuning of MIMO systems based on simultaneous perturbation stochastic approximation," *Expert Syst. Appl.*, vol. 41, no. 14, pp. 6361–6370, 2014.

[51] A. Yonezawa, H. Yonezawa, and I. Kajiwara, "Efficient parameter tuning to enhance practicability of a model-free vibration controller based on a virtual controlled object," *Mech. Syst. Signal Process.*, vol. 200, p. 110526, 2023.

[52] O. Granichin, V. Erofeeva, Y. Ivanskiy, and Y. Jiang, "Simultaneous perturbation stochastic approximation-based consensus for tracking

under unknown-but-bounded disturbances," *IEEE Trans. Automat. Contr.*, vol. 66, no. 8, pp. 3710–3717, Aug. 2021.

[53] Y. Tanaka, S. I. Azuma, and T. Sugie, "Simultaneous perturbation stochastic approximation with norm-limited update vector," *Asian J. Control*, vol. 17, no. 6, pp. 2083–2090, 2015.

[54] K. Onozato and Y. Maeda, "Learning of inverse-dynamics and inverse-kinematics for two-link SCARA robot using neural networks," in *SICE Annual Conference 2007*, Sep. 2007, pp. 1031–1034.

[55] Y. Maeda and R. J. P. De Figueiredo, "Learning rules for neuro-controller via simultaneous perturbation," *IEEE Trans. Neural Networks*, vol. 8, no. 5, pp. 1119–1130, Sep. 1997.

[56] J. L. Maryak and D. C. Chin, "Global random optimization by simultaneous perturbation stochastic approximation," *IEEE Trans. Automat. Contr.*, vol. 53, no. 3, pp. 780–783, 2008.

[57] F. L. Lewis and D. Vrabie, "Reinforcement learning and adaptive dynamic programming for feedback control," *IEEE Circuits Syst. Mag.*, vol. 9, no. 3, pp. 32–50, 2009.

[58] F. Y. Wang, H. Zhang, and D. Liu, "Adaptive dynamic programming: An introduction," *IEEE Comput. Intell. Mag.*, vol. 4, no. 2, pp. 39–47, May 2009.

[59] D. Liu, S. Xue, B. Zhao, B. Luo, and Q. Wei, "Adaptive dynamic programming for control: A survey and recent advances," *IEEE Trans. Syst. Man, Cybern. Syst.*, vol. 51, no. 1, pp. 142–160, Jan. 2021.

[60] Y. Ouyang, L. Dong, Y. Wei, and C. Sun, "Neural network based tracking control for an elastic joint robot with input constraint via actor-critic design," *Neurocomputing*, vol. 409, pp. 286–295, Oct. 2020.

[61] Y. Ouyang, C. Sun, and L. Dong, "Actor–critic learning based coordinated control for a dual-arm robot with prescribed performance and unknown backlash-like hysteresis," *ISA Trans.*, vol. 126, pp. 1–13, Jul. 2022.

[62] X. Yuan, L. Dong, and C. Sun, "Solver–critic: A reinforcement learning method for discrete-time-constrained-input systems," *IEEE Trans. Cybern.*, vol. 51, no. 11, pp. 5619–5630, Nov. 2021.

[63] A. Malik, Y. Lischuk, T. Henderson, and R. Prazenica, "A deep reinforcement-learning approach for inverse kinematics solution of a high degree of freedom robotic manipulator," *Robotics*, vol. 11, no. 2, p. 44, Apr. 2022.

[64] J. C. Spall, "Feedback and weighting mechanisms for improving Jacobian estimates in the adaptive simultaneous perturbation algorithm," *IEEE Trans. Automat. Contr.*, vol. 54, no. 6, pp. 1216–1229, Jun. 2009.

[65] J. Zhu, L. Wang, and J. C. Spall, "Efficient implementation of second-order stochastic approximation algorithms in high-dimensional problems," *IEEE Trans. Neural Networks Learn. Syst.*, vol. 31, no. 8, pp. 3087–3099, Aug. 2020.

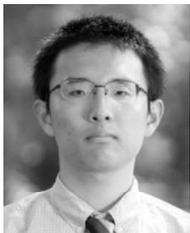
**Ansei Yonezawa** (Member, IEEE) received the B.S., M.S., and Ph.D. degrees in engineering from Hokkaido University, Japan, in 2019, 2021, and 2023, respectively.

Since 2023, he has been a specially appointed Assistant Professor with the Graduate School of Engineering, Hokkaido University. His research interests include active vibration control, robust control, stochastic optimization, fractional-order control, and data-driven control.

He received the Hatakeyama Prize and the Miura Prize given for best students in the field of mechanical engineering from the Japan Society of Mechanical Engineers, in 2019 and 2021, respectively.

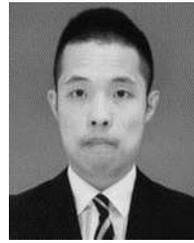
**Heisei Yonezawa** (Member, IEEE) received the B.S., M.S., and Ph.D. degrees in engineering from Hokkaido University, Japan, in 2017, 2019, and 2021, respectively.

Since 2021, he has been an Assistant Professor with the Graduate School of Engineering, Hokkaido University. His research interests include powertrain, vehicle systems, vibration control, robust control, and optimization.

He received the Miura Prize for best students in the field of mechanical engineering from the Japan Society of Mechanical Engineers, in 2019.

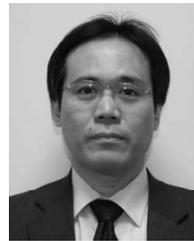
**Itsuro Kajiwara** received the B.S. degree in engineering from Tokyo Metropolitan University, in 1986, and the M.S. and Ph.D. degrees in engineering from the Tokyo Institute of Technology, in 1988 and 1993, respectively.

From 1990 to 2000, he was an Assistant Professor with the School of Engineering, Tokyo Institute of Technology, where he was an Associate Professor with the Graduate School of Engineering, from 2000 to 2008. Since 2009, he has been a Professor with the Graduate School of Engineering, Hokkaido University. His research interests include vibration, control, structural health monitoring, and laser application.